%% file: root.tex
\documentclass[a4paper, 10pt, conference]{ieeeconf}
\usepackage[utf8]{inputenc}
\IEEEoverridecommandlockouts                              %

\usepackage[pass]{geometry}
\usepackage{amsmath,amssymb,amsfonts}
\usepackage{mathtools}                              %
\usepackage{algorithmic}
\usepackage{graphicx}
\usepackage{textcomp}
\usepackage{xcolor}
\usepackage{placeins}
\usepackage{caption}
\usepackage{subcaption}
\usepackage{dsfont}
\usepackage{hyperref}
\usepackage{MnSymbol}
\usepackage{booktabs}
\usepackage{tabularx}

\usepackage[backend=bibtex,bibstyle=ieee,citestyle=numeric-comp,maxbibnames=3,maxcitenames=3,doi=false,url=false,eprint=false]{biblatex}

\usepackage[nameinlink]{cleveref}
\Crefname{equation}{Eq.}{Eq.}
\Crefname{figure}{Fig.}{Fig.}
\Crefname{tabular}{Tab.}{Tab.}
\Crefname{section}{Sec.}{Sec.}

\DeclareRobustCommand{\todo}[1]{}

\newcommand{\nit}[1]{\mathrm{#1}}

\DeclareRobustCommand{\pnidx}[1]{{}_{\nit{#1}}}

\DeclareRobustCommand{\vect}[1]{\boldsymbol{#1}}

\DeclareRobustCommand{\EX}{\mathop{\mathbb{E}}\limits}

\begin{document}

\title{Can You See Me Now?\\ Blind Spot Estimation for Autonomous Vehicles using Scenario-Based Simulation with Random Reference Sensors}
\author{Marc Uecker\textsuperscript{1} and J. Marius Zöllner\textsuperscript{1,2}%
\thanks{\textsuperscript{1}\,Department of Technical Cognitive Systems, FZI Research Center for}%
\thanks{\phantom{\textsuperscript{1}\,}Information Technology, Germany.
	{\tt\small \{surname\}@fzi.de}}
\thanks{\textsuperscript{2}\,Karlsruhe Institute of Technology (KIT), Germany.}%
}
\maketitle

\input{sections/00_abstract.tex}

\input{sections/01a_intro.tex}

\input{sections/01b_related_work.tex}

\input{sections/02_methods.tex}

\input{sections/04_experiments.tex}

\input{sections/03_impl_details.tex}

\input{sections/05_conclusion.tex}

\printbibliography
\vspace{12pt}
\end{document}

%% file: sections/00_abstract.tex
\begin{abstract}
In this paper, we introduce a method for estimating blind spots for sensor setups of autonomous or automated vehicles and/or robotics applications.
In comparison to previous methods that rely on geometric approximations, our presented approach provides more realistic coverage estimates by utilizing accurate and detailed 3D simulation environments.
Our method leverages point clouds from LiDAR sensors or camera depth images from high-fidelity simulations of target scenarios to provide accurate and actionable visibility estimates.
A Monte Carlo-based reference sensor simulation enables us to accurately estimate blind spot size as a metric of coverage, as well as detection probabilities of objects at arbitrary positions.
\end{abstract}

%% file: sections/01a_intro.tex
\section{Introduction \todo{1.5 pages}}
\label{sec:intro}

With the rise of higher automation levels in autonomous driving, more and more driving responsibility is taken away from human drivers, and placed on assistance systems and automated or autonomous driving functions.
However, the software components for these driving functions rely on sufficient coverage from the underlying sensor data in order to function correctly.
After all, even sophisticated tracking and prediction algorithms can only provide limited insight on the vehicle's environment beyond what the sensors are able to perceive.
Therefore, the placement and coverage of the sensor setup has grave and direct implications on the safety of a driving system.
Any potential blind spots could lead to accidents, which may result in injuries to pedestrians, passengers, or damage to the vehicle and surrounding infrastructure \cite{bulumelleSimulatingImpactBlindspots2015}.
For this reason, the sensor setup of autonomous vehicles, prototypes, and vehicles with advanced driver assistance systems must be carefully considered to ensure safe and efficient operation.

\begin{figure}[!t]
\begin{center}
\captionsetup{justification=centering}
    \begin{subfigure}{0.6\linewidth}
    \includegraphics[trim={0cm 0cm 0cm 0cm},clip,width=\linewidth]{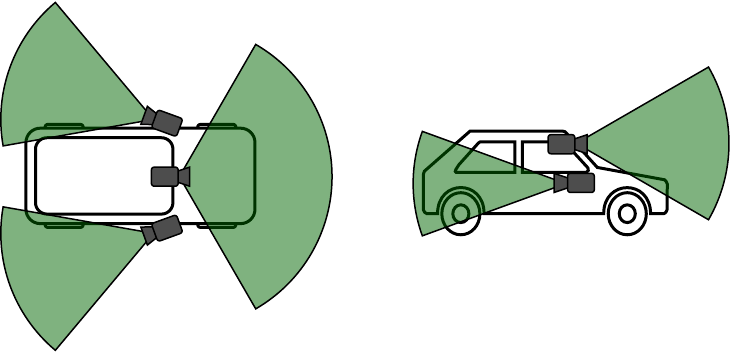}%
    \caption{Sensor setup with three cameras}%
    \label{fig:page1:a}
    \end{subfigure}%
    \vspace{0.01\linewidth}
    \begin{subfigure}{1.0\linewidth}%
    \includegraphics[width=\linewidth]{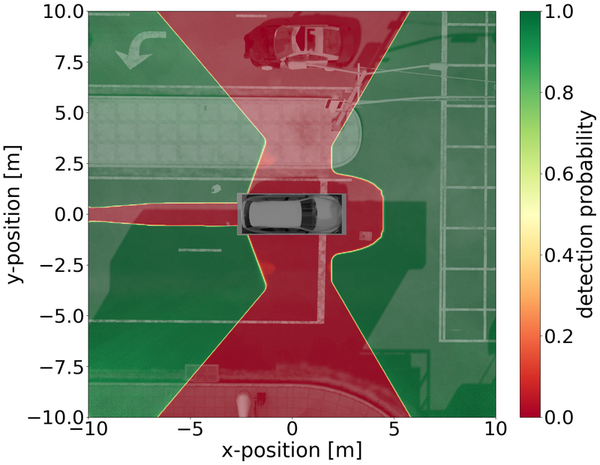}%
    \caption{Detection probability of a small obstacle (0.4~m radius) at ground level.}%
    \label{fig:page1:b}
    \end{subfigure}%
    \vspace{0.01\linewidth}
    \begin{subfigure}{1.0\linewidth}%
    \begin{center}%
    \includegraphics[width=0.6\linewidth]{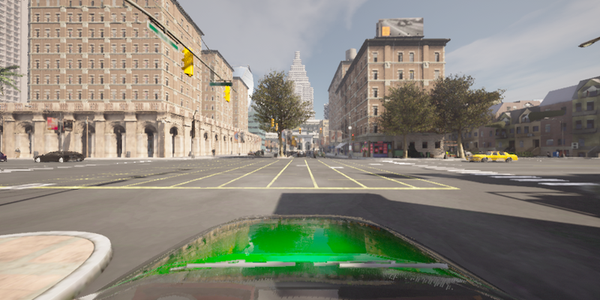}%
    \vspace{0.01\linewidth}
    \includegraphics[width=0.495\linewidth]{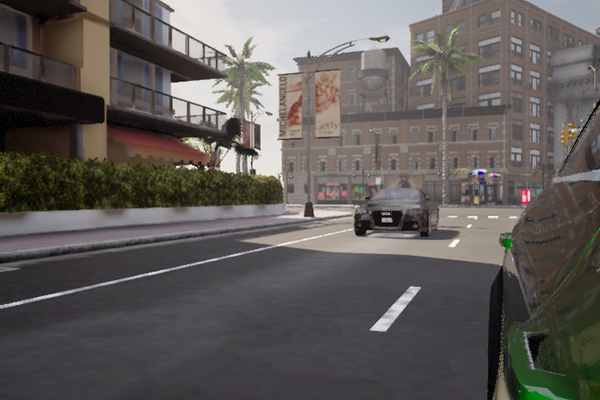}%
    \hfill%
    \includegraphics[width=0.495\linewidth]{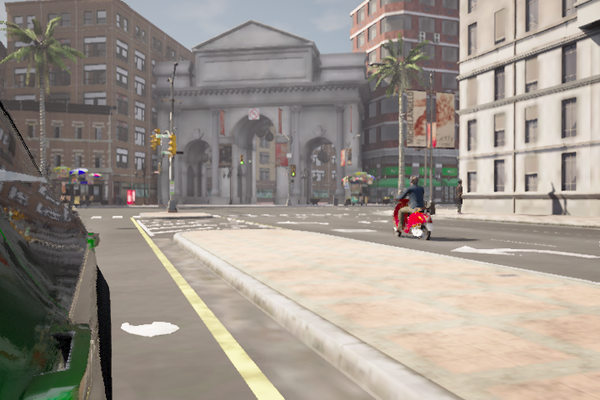}%
    \caption{Camera images from the simulation}%
    \label{fig:page1:c}
    \end{center}%
    \end{subfigure}%
\end{center}%
\vspace{-8pt}%
\caption{A camera setup with one front and two rear view cameras is simulated.
The resulting coverage map \hyperref[fig:page1:b]{(b)} provides an accurate estimate of the vehicle's field of view.}
\label{fig:page1}
\end{figure}
\begin{figure*}[!t]
    \captionsetup[subfigure]{justification=centering}
    \newsavebox{\mybox}%
    \savebox{\mybox}{\includegraphics[width=0.245\linewidth]{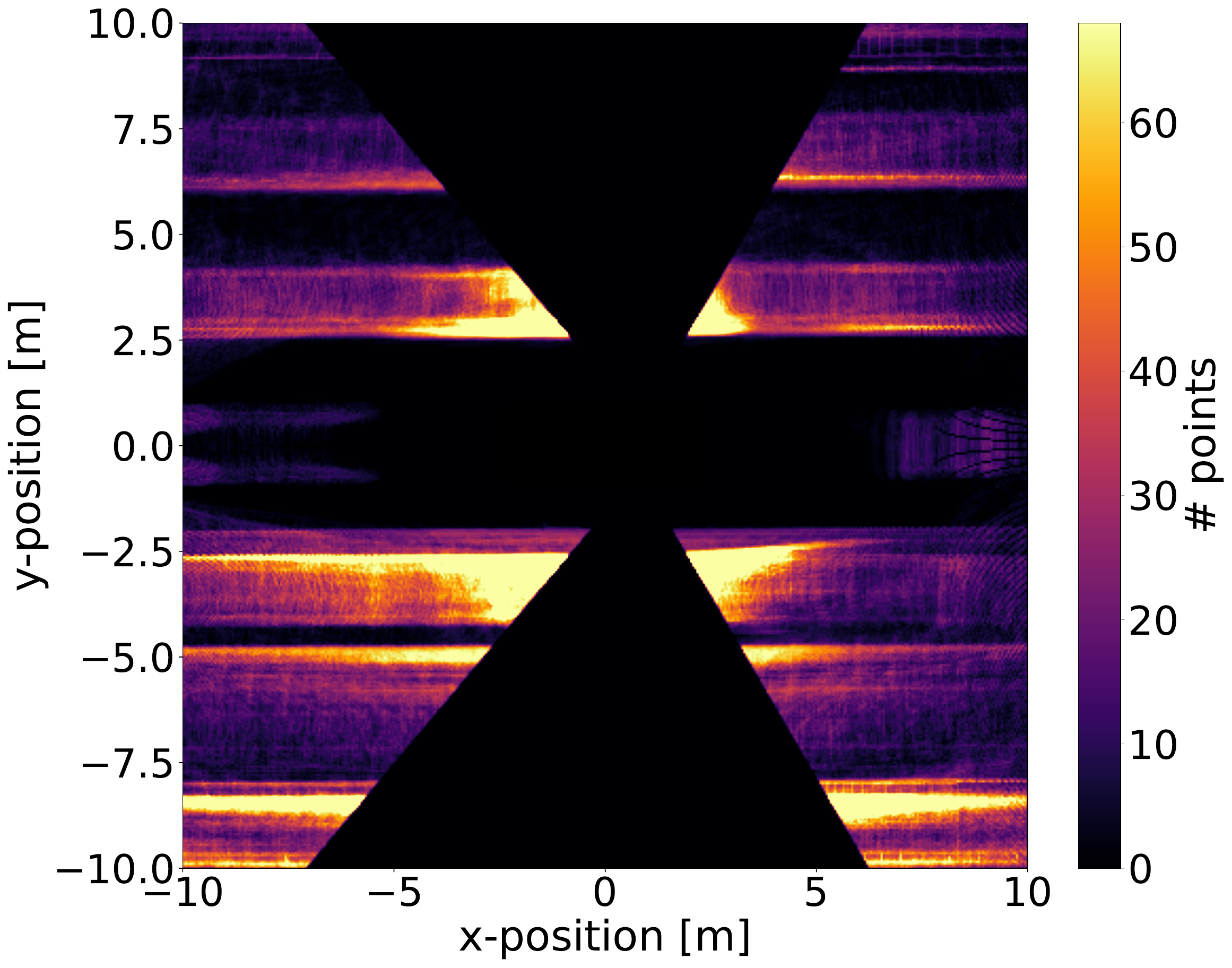}}%
    \begin{subfigure}[t]{0.245\linewidth}%
    \resizebox{!}{\ht\mybox}{\includegraphics[width=\linewidth]{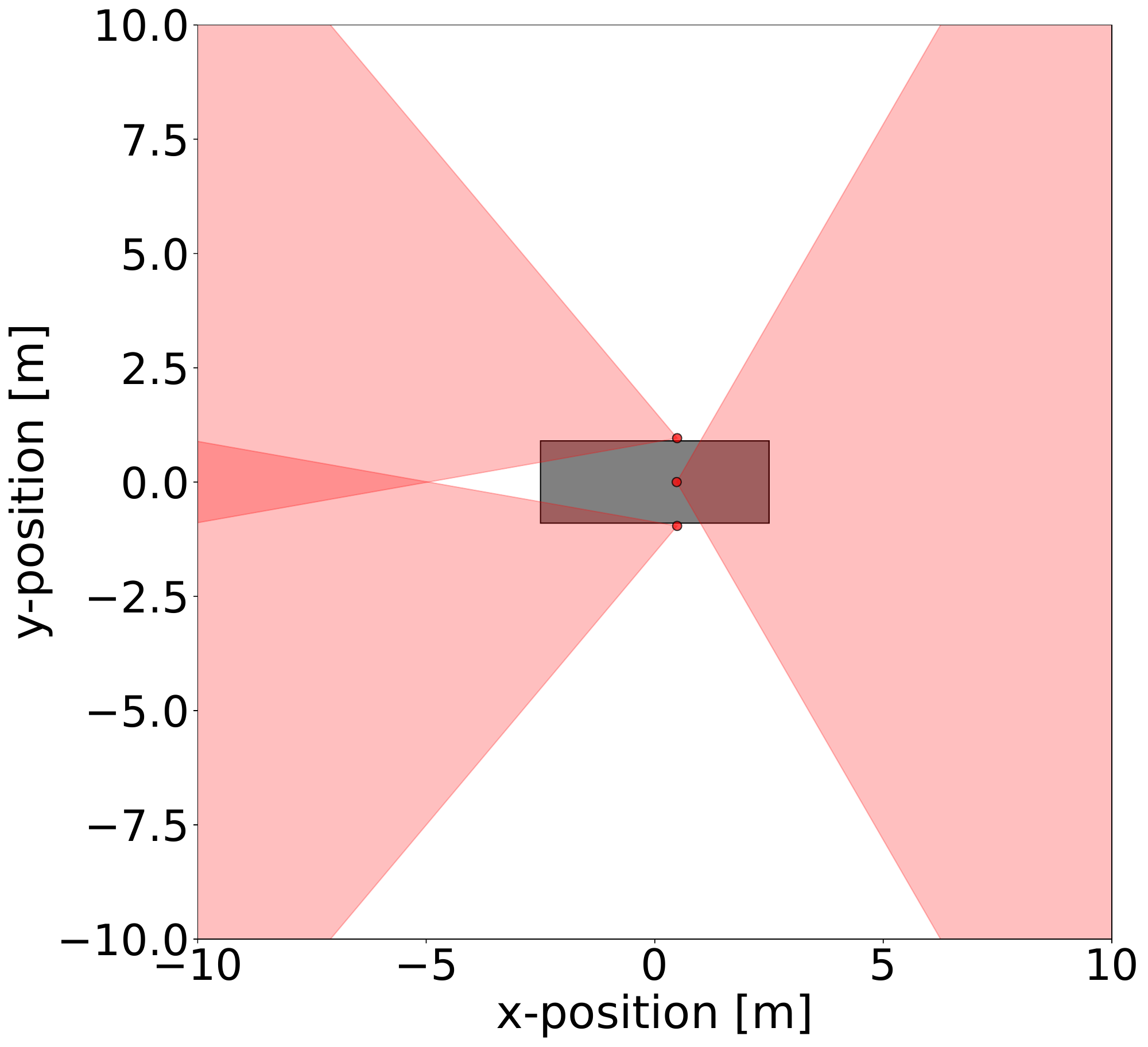}}%
    \caption{Geometric approximation\\(2D)\label{fig:related:geo}}%
    \end{subfigure}%
    \hfill
    \begin{subfigure}[t]{0.245\linewidth}%
    \includegraphics[width=\linewidth]{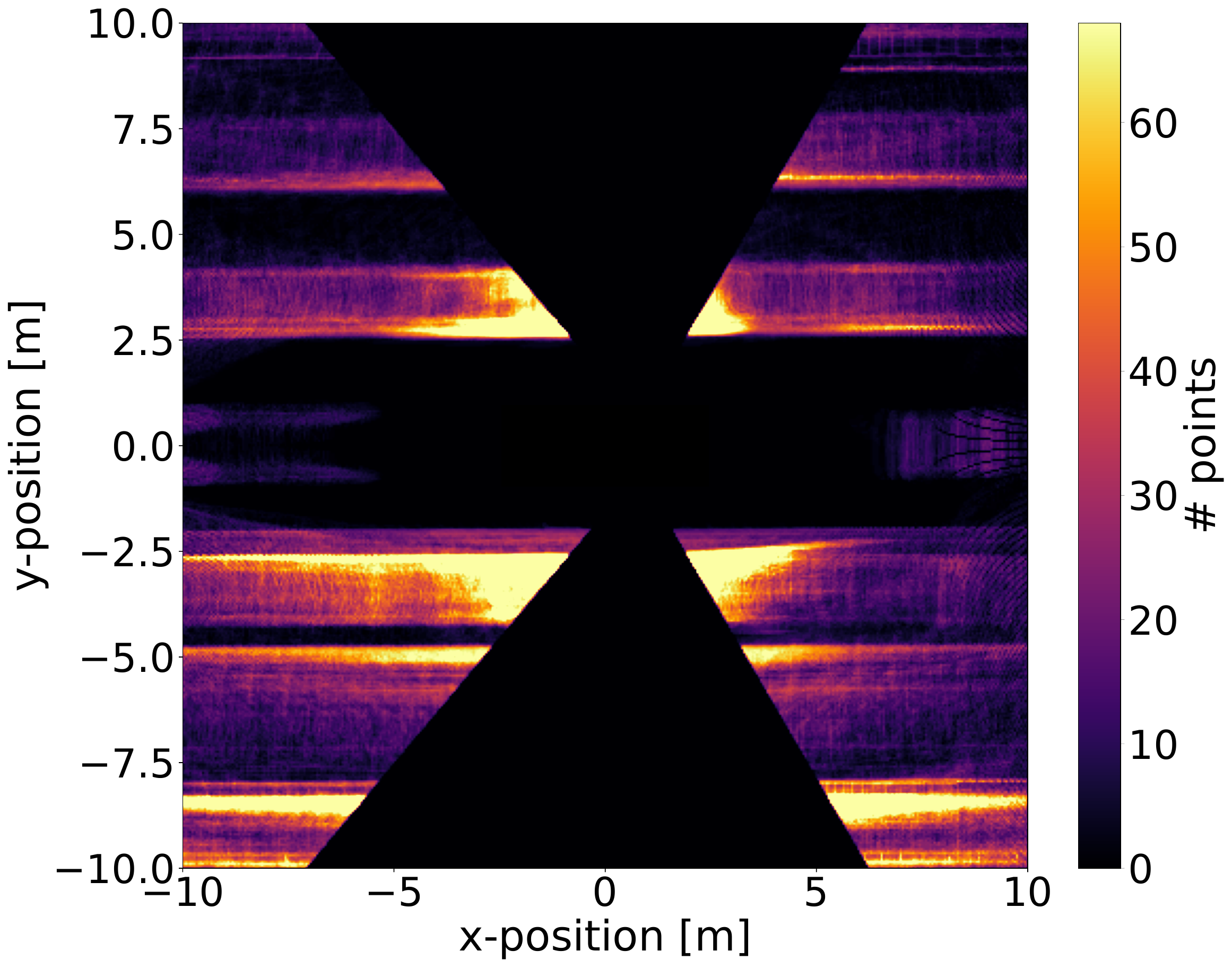}%
    \caption{Point distribution histogram\\(obstacle level)\label{fig:related:hist}}%
    \end{subfigure}%
    \hfill
    \begin{subfigure}[t]{0.245\linewidth}%
    \includegraphics[width=\linewidth]{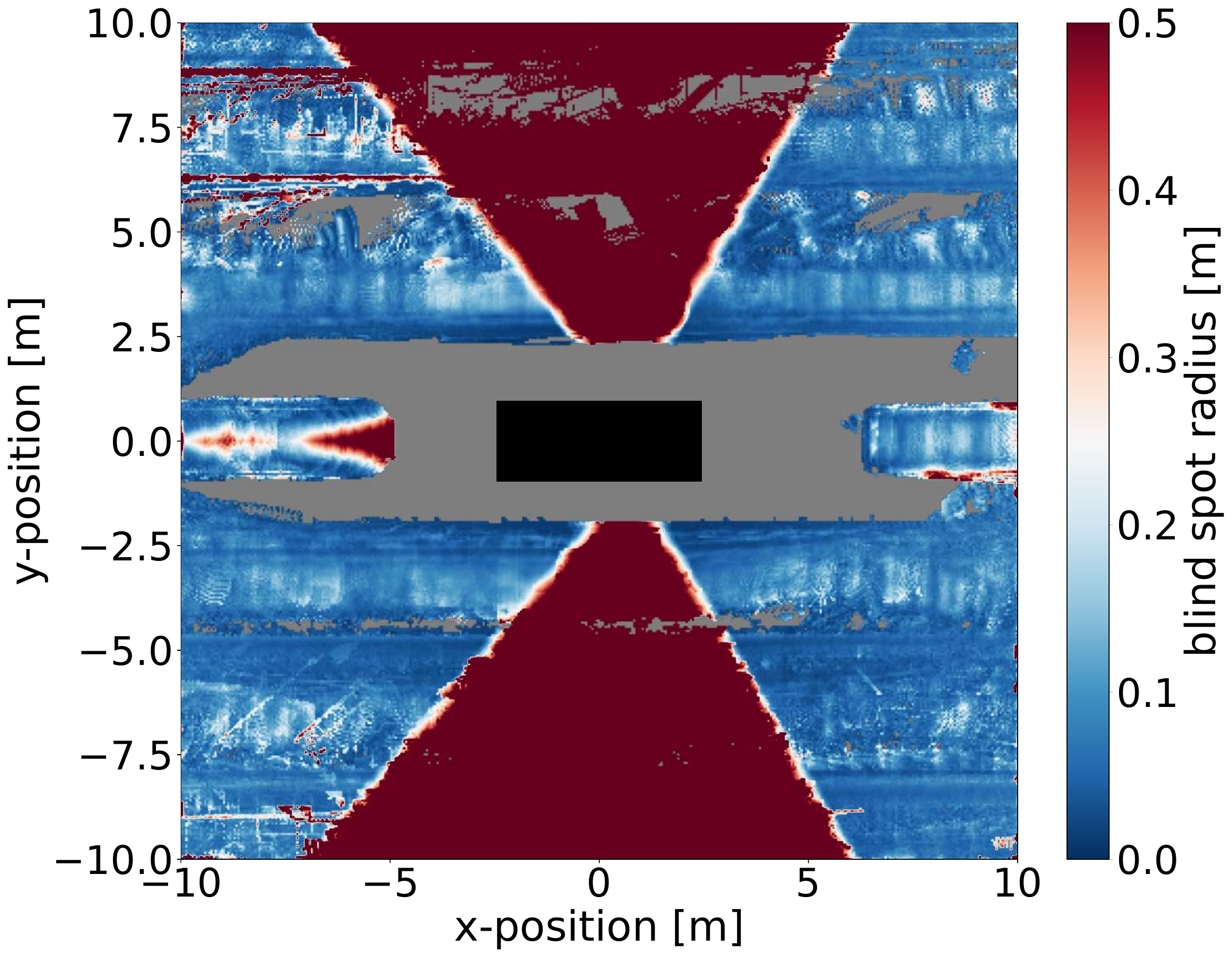}%
    \caption{Our method\\(obstacle level)\label{fig:related:our_obst}}%
    \end{subfigure}%
    \hfill
    \begin{subfigure}[t]{0.245\linewidth}%
    \includegraphics[width=\linewidth]{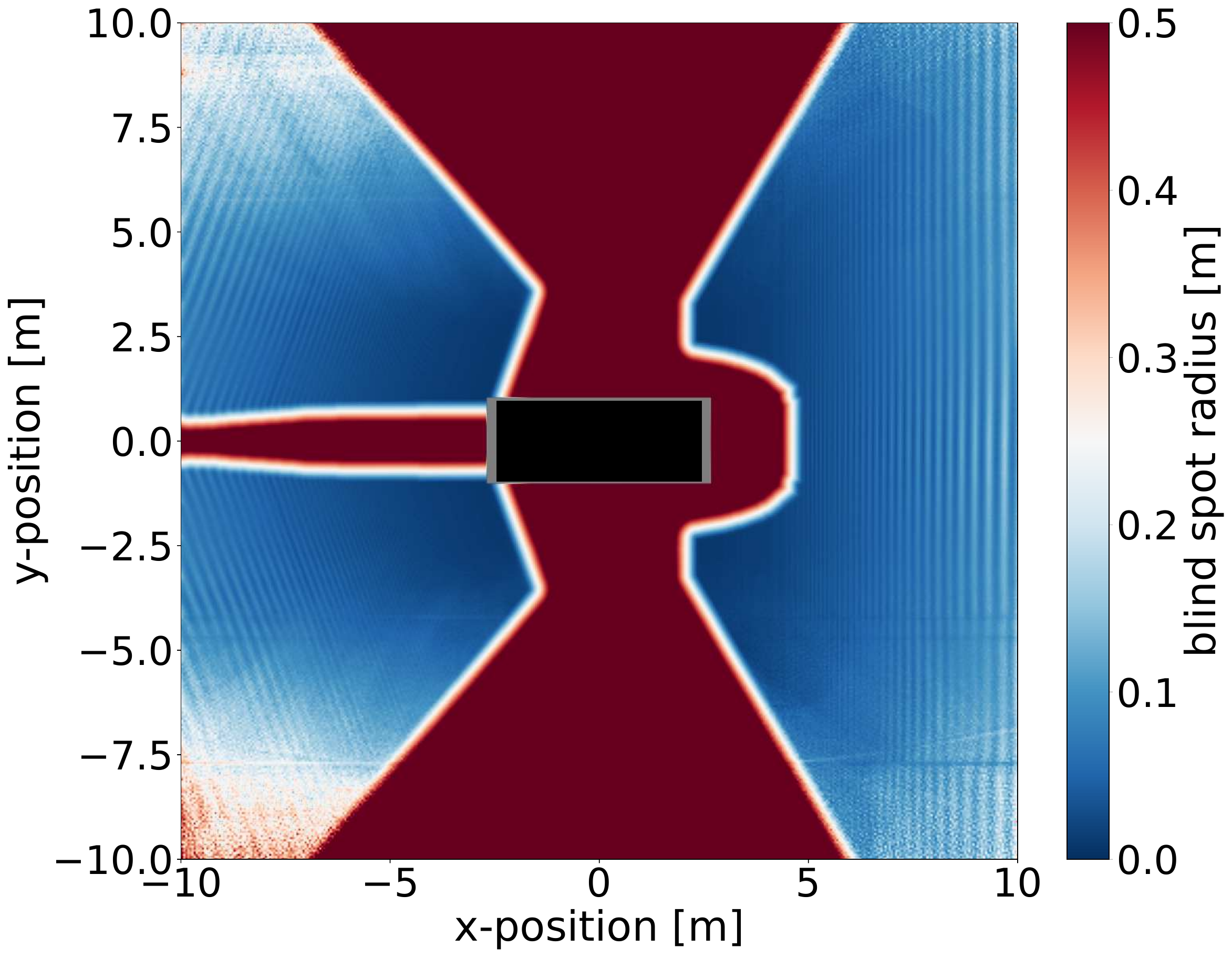}%
    \caption{Our method\\(ground level)\label{fig:related:our_ground}}%
    \end{subfigure}%
    \caption{Comparing our method to common methods of illustrating sensor setup coverage. The underlying camera setup is the same as illustrated in \Cref{fig:page1}.\label{fig:related}}%
\end{figure*}
Potential candidates for sensor setups therefore need be inspected for blind spots and sufficient coverage of their surroundings.
Simple geometric approximations for these estimates have existed for a long time.
In their simplest form, they are easy to implement for single-sensor and even multi-sensor setups.
However, they necessarily abstract away many details of the underlying physical system.
In a simplified model, omitting certain effects can lead to reduced realism and undiscovered failure modes in real-world use cases.
Using geometric approximations, it is difficult to accurately model complex geometry, occlusion, and sensor-specific characteristics, such as limited range and finite sensor resolution.
As sensor setups become more involved, incorporating LiDAR- and camera-based sensing into automated driving functions, the need for accurate modeling of their weaknesses becomes even more crucial.
Advanced simulations can assist in refining these approximations, providing more realistic sensor data.
However, there is as of yet a limited body of research devoted to accurate, yet simple, and human-interpretable methods of visualizing and interpreting this high-fidelity simulation data to estimate and maximize sensor coverage.

In this work, we propose a novel method to analyze the coverage of LiDAR and camera sensors in realistic and application-specific scenarios.
Our method generalizes across LiDAR and camera sensors via a unified pointcloud data format, which is easily computed from simulation data.
Based on this data, we introduce two related local coverage metrics, which measure a sensor setup's blind spots and detection quality. 
Through rasterized visualizations, our presented metrics allow quick identification and localization of low-coverage regions or blind spots.
Aggregation across regions of interest provides simple and interpretable metrics by which different sensor setup candidates can be compared. 

We utilize a Monte Carlo-based reference sensor in addition to high-fidelity simulation in order to estimate blind spots more accurately than previous methods.
Our approach is easily applied to different sensor setups, and scenarios, as well as able to accurately handle complex geometries up to the level of accuracy provided by the underlying simulation engine.
Our method can be applied on top of existing autonomous vehicle simulation platforms such as CARLA~\cite{dosovitskiyCARLAOpenUrban2017}, dSpace AURELION, or IPG CarMaker.
Therefore, it can easily be integrated into existing scenario-based validation toolchains for Software-in-the-Loop (SIL) validation, or used to evaluate and optimize conceptual designs for sensor setups in early stages of vehicle development.
We publish our coverage estimation and visualization code at \href{https://github.com/Pyrestone/CanYouSeeMeNow}{\url{github.com/Pyrestone/CanYouSeeMeNow}}.

In the following sections, we provide an overview of existing visibility estimation methods, and describe our approach in greater detail, before presenting some experiments, which showcase the versatility of our method.

%% file: sections/01b_related_work.tex
\section{Related Work \todo{0.5 pages}}
\label{sec:related_work}

\begin{figure*}[t]
    \centering
    \includegraphics[width=\textwidth]{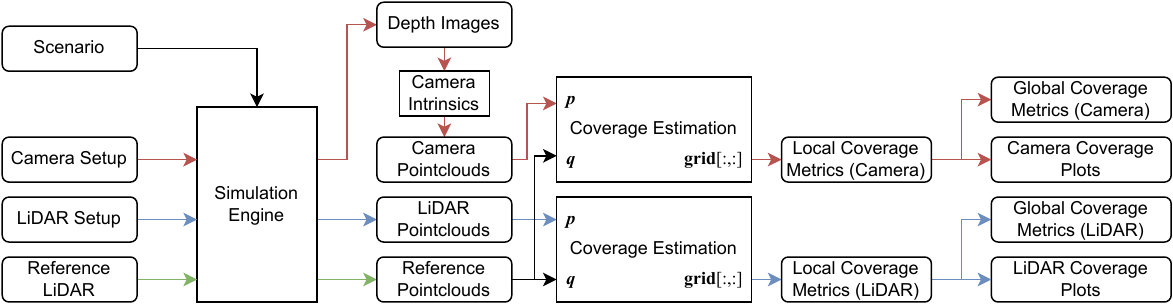}
    \caption{An overview of the presented method: Sensor data from either camera or LiDAR sensors is compared to a reference pointcloud to estimate and illustrated the coverage of different sensor setups.}
    \label{fig:method_overview}
\end{figure*}

In classical computer vision, the problem of placing sensors for maximum visibility is known as the \emph{Art Gallery Problem}~\cite{orourkeArtGalleryTheorems1987}.
In its mathematical formulation, the art gallery problem is a mathematical optimization task formulation, where a static, analytically-defined environment or target geometry, as well as sensor parameters are known in advance.
The objective of the art gallery problem is to determine the minimum number of sensors required and their optimal locations to ensure complete or maximized coverage and visibility of an entire environment / target geometry~\cite{orourkeArtGalleryTheorems1987}.
This problem has applications in surveillance, robotics, and computer graphics, where efficient sensor placement is crucial for monitoring and capturing information.

Various extensions of the basic geometric optimization formulation have been studied throughout the years, extending the scope of sensor placement problems.
Many extensions explore additional constraints and considerations, such as depth of field~\cite{cowanAutomaticSensorPlacement1988}, additional lighting requirements~\cite{yiOptimalSensorLight1995}, limited sensor range, or occlusions within the target geometry~\cite{tarabanisComputingOcclusionfreeViewpoints1996, reedConstraintbasedSensorPlanning2000}.

Solutions and approaches for the art gallery problem have also been applied to applications related to automated vehicle sensors~\cite{mittalVisibilityAnalysisSensor2004}, roadside infrastructure~\cite{vijayOptimalPlacementRoadside2021}, and other robotics tasks~\cite{chenAutomaticSensorPlacement2004}.

In the domains involving mobile systems, such as sensor-equipped vehicles, the requirement of a known and stationary environment is no longer fulfilled, which significantly increases the difficulty compared to the classic art gallery problem.
Lack of knowledge about the target geometry limits many approaches to approximating the field of view of a sensor as a simple circular sector or viewing frustum, as illustrated in \Cref{fig:related:geo}.
This is also the level of detail provided in industry-standard tools for designing sensor setups~\cite{niuObjectDetectionTracking2018}.
This geometric approximation greatly simplifies computational and geometric reasoning, and may be justified in order to make numerical optimization computationally feasible.
However, the resulting visibility estimates provide very limited information about the performance of a sensor setup in its typical environment or usage scenario.
Considerations regarding sensor parameters (e.g. resolution, vertical placement and field-of-view), and susceptibility to environment occlusions also can't always be adequately addressed using these geometric approximations.

To address the limitations of geometric formulations, multiple methods have instead utilized sensor simulations to compute estimates of sensor setup coverage~\cite{deyVESPAFrameworkOptimizing2021,chitlangiaImprovingPerceptionSensor2021, karleEDGARAutonomousDriving2023}.
However, even when using simulations, most methods still use highly simplified target geometry, or static placeholder environments to estimate their sensor setup coverage outside a static ego vehicle or environment model~\cite{deyVESPAFrameworkOptimizing2021,karleEDGARAutonomousDriving2023,vijayOptimalPlacementRoadside2021}
Otherwise, the pursuit of increased realism in simulations also introduces a higher level of complexity.
For instance, \Cref{fig:related:hist} show a histogram of sensor detections aggregated across multiple time-steps of a simulated dynamic scenario.
The underlying point cloud distribution presented in this illustration contains entangled information about the target geometry and the coverage of the sensor setup.
As illustrated by this example, interpreting simulation data without appropriate aggregation and visualization methods is often not straight-forward. 
As a result, aggregation of simulation data into metrics which can be compared and optimized, or insightful visualizations for human decision-making is often required~\cite{borgerSensorVisibilityEstimation2022a}.

In the space of coverage metrics for automated optimization from a dynamic traffic scenario, the metric presented in~\cite{chitlangiaImprovingPerceptionSensor2021} is perhaps the most elaborate method in the available literature.
They employ an advanced information-theoretic metric based on extensive simulation data in order to provide a surrogate metrics for LiDAR sensor setup coverage.
Their results show that -- at least in their simulation -- optimized sensor setups with high visibility also increase the performance of machine-learning based tasks such as 3D object detection.
However, the computational cost of high-fidelity simulation, combined with algorithmic optimization which require many iterative simulations to converge represents a significant barrier to entry.
Similarly, many other approaches apply genetic algorithms, random search, or other general-purpose optimization methods to the coverage metrics or cost functions which they define~\cite{chenAutomaticSensorPlacement2004, dybedalOptimalPlacement3D2017, huInvestigatingImpactMultilidar2022, kimPlacementOptimizationMultiple2020, reedConstraintbasedSensorPlanning2000, vijayOptimalPlacementRoadside2021}.

In practice, however, the design process of sensor setups is often driven more by manual design, with simulations only providing insight for human decision-making~\cite{karleEDGARAutonomousDriving2023}.
This is often necessary, as there are many hard and soft constraints on what is considered \emph{reasonable} sensor placement.
Many of these (e.g., regulatory compliance, aesthetics, availability of mounting hardware) would be tedious or difficult to explicitly encode into a cost function for optimization.
Therefore it is often easier to compare multiple human-selected designs.
Humans have also been shown to be exceedingly efficient at finding near-optimal solution candidates for computationally difficult optimizations problems when given the right insights into the problem~\cite{hidalgo-herreroComparingProblemSolving2013,holzingerGlassboxInteractiveMachine2017}.
This is also the use case we focus on in this paper, as automated optimization would require running many instances of our ray-tracing based simulation, which exceeds our computational resources.
Therefore, our method focuses on providing actionable insights into the visibility and blind spots of candidate sensor setups, as well as providing simple-to-understand coverage metrics by which different sensor setup candidates can be compared.

%% file: sections/02_methods.tex
\section{Methods \todo{1.5 pages}}
\label{sec:methods}
A high-level overview of our method is provided in \Cref{fig:method_overview}.
To evaluate the coverage of a sensor setup at a particular location, we interpret both cameras and LiDAR sensors as perceiving 3D pointclouds.
\Cref{sec:methods:pointclouds} explains how we transform camera depth images to pointclouds.
In addition to the vehicle's sensors, we utilize an additional pointcloud from what we refer to as the \emph{reference sensor}, to probe the coverage of the vehicle's sensor setup.
\Cref{sec:methods:reference} goes into further detail on how the reference sensor is set up, and why it is necessary.
\Cref{sec:method:metric} describes the formula and geometric interpretation of our blind spot radius metric.
Finally, \Cref{sec:method:viz} describes how we produce two kinds of visualizations from the sparsely sampled metric.

\subsection{Sensor Data as Point Clouds}
\label{sec:methods:pointclouds}
In order to accurately estimate a three-dimensional field of view for sensors, we first obtain three-dimensional pointclouds, i.e. sets of points ($\pnidx{veh}\vect{p}$), for our simulated sensors in the vehicle coordinate frame.
This is common practice for LiDAR sensors, but is rarely done for monocular camera data.
This step is straightforward for simulated LiDAR sensors, as they directly provide 3D points, and the only transformation necessary is a coordinate system transformation to the vehicle coordinate system.
With a known homogeneous transformation matrix ($\pnidx{veh}T\pnidx{sensor}$) from a calibrated sensor setup, this step is a simple matrix multiplication:
\begin{align}
\label{eqn:extrinsic}
\pnidx{veh}\vect{p} = \pnidx{veh}T_\nit{sensor} \cdot \pnidx{sensor}\vect{p}
\end{align}

For camera sensors, this step is slightly more involved, as there often is no appropriate camera sensor model in simulation engines which directly provides 3D pointclouds.
There is however a depth camera, which provides the depth ($\pnidx{cam}\vect{p}_z$) of the 3D point ($\vect{p}$) captured at each pixel coordinate $\pnidx{px}\vect{p}$ from the camera's coordinate origin.
\begin{align}
\label{eqn:perspective_projection}
\begin{pmatrix}
\pnidx{lens}p_x\\
\pnidx{lens}p_y
\end{pmatrix} &=& 
\begin{pmatrix}
\pnidx{cam}p_x/\pnidx{cam}p_z\\
\pnidx{cam}p_y/\pnidx{cam}p_z
\end{pmatrix} \quad |\;\; \pnidx{cam}p_z > 0\\
\label{eqn:lens_distortion_general}
\begin{pmatrix}
\pnidx{img}p_x\\
\pnidx{img}p_y
\end{pmatrix} &=&
\nit{distortion\_model}(\pnidx{lens}\vect{p},\,\vect{k})\\
\label{eqn:intrinsic_camera}
\begin{pmatrix}
\pnidx{px}p_x\\
\pnidx{px}p_y
\end{pmatrix} &=&
\begin{pmatrix}
f_x & 0 & c_x\\
0 & f_y & c_y
\end{pmatrix}
\begin{pmatrix}
\pnidx{img}p_x\\
\pnidx{img}p_y\\
1
\end{pmatrix}
\end{align}
Given an intrinsic calibration of the camera, with known distortion (if any), as provided in \Cref{eqn:perspective_projection,eqn:lens_distortion_general,eqn:intrinsic_camera}, one can reverse the appropriate operations to retrieve the original 3D point. 
This process is formally described in \Cref{eqn:reverse_intrinsic_camera,eqn:reverse_lens_distortion_general,eqn:reverse_perspective_projection} \cite{hartleyMultipleViewGeometry2003}.
\begin{align}
\label{eqn:reverse_intrinsic_camera}
\begin{pmatrix}
\pnidx{img}p_x\\
\pnidx{img}p_y
\end{pmatrix}
&=&
\frac{1}{f_x}
\begin{pmatrix}
\pnidx{px}p_x - c_x\\
\pnidx{px}p_y - c_y
\end{pmatrix}\\
\label{eqn:reverse_lens_distortion_general}
\pnidx{lens}\vect{p}&=&\nit{distortion\_model}^{-1}(\pnidx{img}\vect{p},\,\vect{k})\\
\label{eqn:reverse_perspective_projection}
\begin{pmatrix}
\pnidx{cam}p_x\\
\pnidx{cam}p_y\\
\pnidx{cam}p_z
\end{pmatrix} &=&
\begin{pmatrix}
\pnidx{lens}p_x \cdot \pnidx{cam}p_z\\
\pnidx{lens}p_y \cdot \pnidx{cam}p_z\\
\nit{depth}
\end{pmatrix}
\end{align}
Notably, the pixel coordinates in the lens coordinate system $\pnidx{lens}\vect{p}$ can be stored after being pre-computed once, which significantly decreases the computational overhead of the inverse distortion model calculation.
Finally, the points in the camera coordinate system can be transformed to the vehicle coordinate system analogous to \Cref{eqn:extrinsic}.
In the following, we will handle camera and LiDAR sensors identically, as both data can be interpreted as 3D point clouds.

\begin{figure}[!t]
    \centering
    \includegraphics[width=\linewidth]{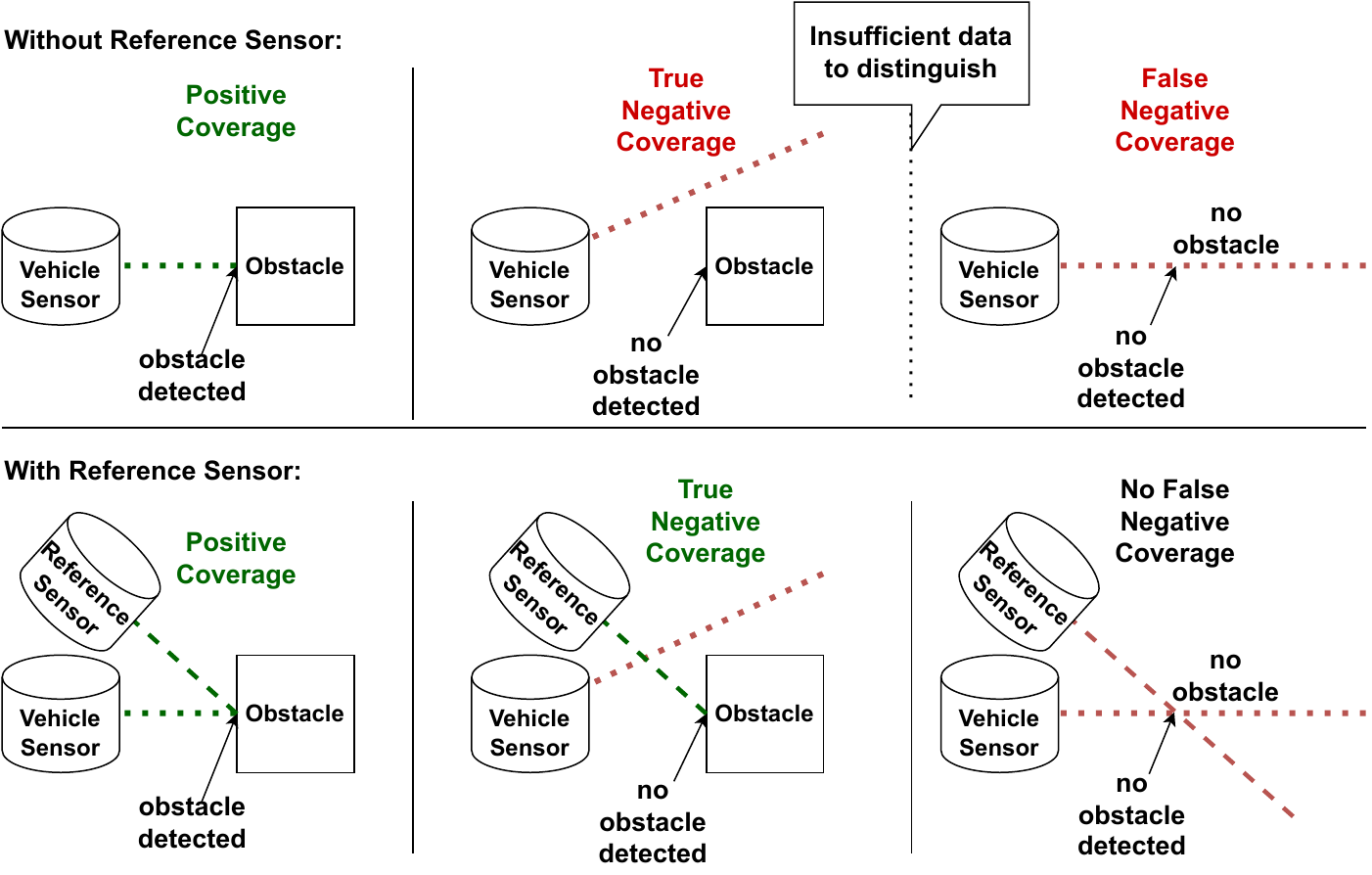}
    \caption{Illustrating the need for a reference sensor:\\Without a reference sensor (top), a lack of sensor detections at a location may either be the result of lacking coverage (center) or missing target geometry (right). Probing the target geometry with a reference sensor (bottom) alleviates this issue.}
    \label{fig:negative_coverage}
\end{figure}

\subsection{Probing with Random Reference Sensor}
\label{sec:methods:reference}
In order to evaluate the coverage of a sensor setup, it is not sufficient to take the 3D pointcloud of the setup and mark all regions with 3D points as covered and all empty regions as blind spots.
The main issue with this approach is that 3D space is mostly devoid of target geometry.
This issue is illustrated in \Cref{fig:negative_coverage} (top row): If the sensor detects a point in a particular location, this can be interpreted as evidence of positive coverage in this location (top left).
However, if the sensor does not detect anything in a particular location, this does not necessarily mean that the sensor has no coverage in this location.
Instead, it may simply mean that there is nothing at this location for the sensor to detect (compare \Cref{fig:negative_coverage} top center vs top right).
To distinguish these two cases, one must refer back to the underlying 3D geometry of the scene and evaluate the coverage only where detectable obstacles are present.
Our proposed solution to this problem is to simulate an additional LiDAR sensor as a reference to scan the geometry of the environment.
As seen in the lower row of \Cref{fig:negative_coverage}, adding this reference sensor allows us to differentiate between positive and negative coverage at a certain location.
Without a reference sensor, a naive method must assume that all regions without sensor data are evidence of lacking coverage.

To avoid visibility issues or scan-pattern related artifacts from probing in fixed positions, we place the reference sensor\footnote{For simplicity, we describe this method with one sensor and one reference sensor, but it can easily be extended to examine the coverage of multiple fused sensors, or multiple reference sensors by fusing the appropriate pointclouds.} in a random position around the vehicle at each time step.
Across multiple simulation steps, this simulates a sensor setup with maximum visibility, where the vehicle is entirely covered in sensors.

\subsection{Coverage Metric}
\label{sec:method:metric}
As described in \Cref{sec:methods:reference}, a reference sensor is necessary to differentiate false negative coverage from true negative coverage.
However, in real simulations, there is almost never an exact match between the locations of the points provided by the vehicle and reference sensor.
Instead, the points are typically at least slightly offset.
Thus, calculating the coverage at a location with a known obstacle becomes a nearest-neighbor matching problem.

\begin{figure}[!b]%
    \centering%
    \includegraphics[width=1.0\linewidth]{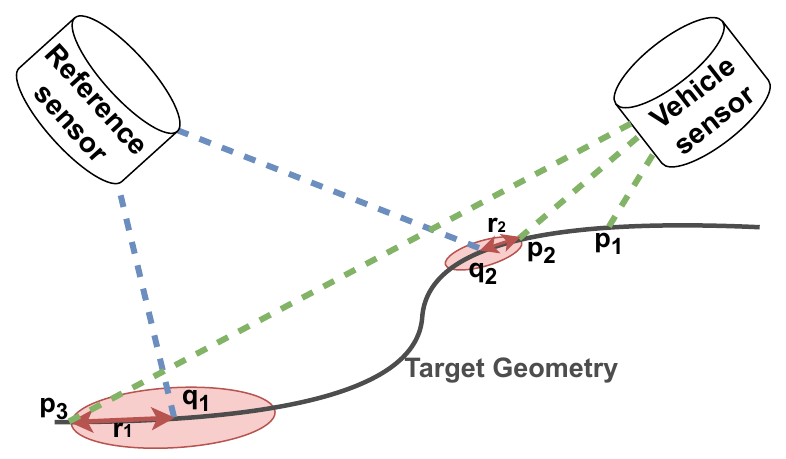}%
    \caption{Computation of the blind spot radius $r_1$ and $r_2$ at the probe points $\vect{q}_1$ and $\vect{q}_2$.}%
    \label{fig:probing_distance}%
\end{figure}

Given two point clouds $P_{\nit{ref},t}$, and $P_{\nit{s},t}$, which respectively come from the reference sensor and the vehicle's combined sensors at time-step $t$, our aim is now to compute a metric which quantifies the visibility at the known target geometry probed by the reference sensor.
To achieve this, for each point $\vect{q}$ from the reference sensor pointcloud, we compute the euclidean distance to to the closest point from the vehicle's sensors under evaluation, as described in \Cref{eqn:distance_metric}.
\begin{align}
\label{eqn:distance_metric}
    r(\vect{q},P_\nit{sensor}) = \min_{\vect{p}_j\in P_\nit{s}}{\left\Vert\vect{q} - \vect{p}_j\right\Vert}
\end{align}
Geometrically, this distance indicates the radius of the largest circle or sphere that would fit inside the blind spot around the probe point $\vect{q}$ from the reference sensor, without containing any of the vehicle sensors' points.
Therefore, the distance metric is an estimate of the blind spot radius at the reference point, as illustrated in \Cref{fig:probing_distance}.

\subsection{Visualization and Aggregation of Metrics}
\label{sec:method:viz}
The samples from our reference sensor data, and thus of the resulting blind spot radius metric, are sparsely scattered across individual points in time and space.
To visualize and analyze these sparse samples, we quantize their position into a 2D grid, and aggregate the grid values across all simulation time steps\footnote{We use 4096 time steps for the simulations in this paper.} to produce a dense 2D representation.
This binning is mathematically described as follows:
\begin{align}
\label{eqn:geometry_binning}
\mathrm{grid}[x,y]=\EX_{t}\left[ \EX_{\vect{q}\in P_{\nit{ref},t}}\left[ r(\vect{q},P_{\nit{s},t})\,\,|\,\,\vect{q}\in \text{cell } (x,y) \right]\right] 
\end{align}
Here, $\EX_{t}\left[\cdot\right]$ is the mean across all simulation time-steps, and $\EX_{\vect{q}\in P_{\nit{ref},t}}\left[\cdot\,|\,\,\vect{q}\in \text{cell } (x,y)\right]$ is the mean across all reference points $\vect{q}$ from the reference sensor pointcloud which lie inside the binning cell at location $(x,y)$.
In this format, the value at each grid location is an average of the values $r$ across all query points $\vect{q}$ that lie within said grid square.
Examples for the resulting sampling grids can be seen in \cref{fig:related:our_ground,fig:related:our_obst,fig:would_you_see_a_child}
Naturally, one could use different aggregation methods for this visualization, such as the maximum across all samples for an estimated worst-case analysis.
This tends to favor outliers such as temporarily occluded areas, which may or may not serve the intended purpose.

If  the detection probability of an object of known size is desired, this can be approximated by thresholding the blind spot radius metric and empirically estimating a probability as the ratio of successful detections.
The following formula represents this process:
\resizebox{\linewidth}{!}{\parbox{1.18\linewidth}{
\begin{align}
\label{eqn:geometry_binning_prob}
\mathrm{grid}[x,y]=\EX_{t}\left[ \EX_{\vect{q}\in P_{\nit{ref},t}}\left[ \mathds{1}(r(\vect{q},P_{\nit{s},t})\leq r_\nit{thresh})\,\,|\,\,\vect{q}\in \text{cell } (x,y) \right]\right]
\end{align}
}}
where $\mathds{1}$ is the indicator function.
The results of these grid-mapping illustrations can be seen in \Cref{fig:page1:b}.
For these illustrations, we not only bin in 2D space, but also split regions of interest along the vertical (z) axis of the vehicle (as shown in \Cref{fig:related:our_ground} vs \Cref{fig:related:our_obst}).
This vertical binning permits separate evaluation of coverage at the height of the typical ground plane, as well as in the region where driving obstacles most often occur.

For the final metric values reported in \cref{tab:sensor_positions,tab:sensor_resolutions}, we report the (non-weighted) mean across all non-empty grid values within the region of interest.

%% file: sections/04_experiments.tex
\section{Experiments \todo{1 page}}
\label{sec:experiments}
We implement our method as a proof-of-concept based on CARLA\cite{dosovitskiyCARLAOpenUrban2017} simulation data.
In this section we aim to cover three common use-cases for analyzing the field-of-view of a sensor setup:
\begin{enumerate}
    \item Inspecting a particular sensor setup candidate for blind spots. (\Cref{sec:experiments:camera})
    \item Comparing the visibility and blind spots of multiple candidate sensor positions. (\Cref{sec:experiments:lidar_pos})
    \item Assessing the impact of different sensor specifications, (e.g. resolution) on the setup's coverage. (\Cref{sec:experiments:lidar_res})
\end{enumerate}

\subsection{Inspecting blind spots for a camera setup}
\label{sec:experiments:camera}
The first use case we address is  inspecting a candidate sensor setup for blind spots.
For this purpose, we choose a three-camera setup with one central front-facing and two rear-facing cameras mounted on the side mirrors, as illustrated in \Cref{fig:page1:a,fig:page1:c}.

We apply our method to simulated depth images from this setup, and show the resulting coverage maps, which are further explained here:
\Cref{fig:related:our_ground,fig:related:our_obst} show the resulting blind spot radius maps for this setup.
\Cref{fig:related:our_ground} shows the estimated blind spot radius ($r$ in \Cref{eqn:distance_metric}) at ground level ($z \approx 0$) in a close-range region of interest around the vehicle ($x,y \in [-10~m,\,10~m]$ respectively).
Here, one can see that self-occlusions result in significant blind spots in front of the vehicle's hood, as well as behind the vehicle.

\Cref{fig:related:our_obst} shows the estimated blind spot radius $r$ at a typical obstacle height ($z \in [0.5~m,\,2.0~m]$) in the same close-range region of interest around the vehicle as \Cref{fig:related:our_ground}.
Here, we can see that the visibility of the sensor setup changes significantly across the vertical axis, with smaller blind spots higher up.
These differences are due to the tapered shape of our simulated ego vehicle\footnote{The vehicle model in our simulation is an \emph{Audi eTron} from CARLA's vehicle library.}.

\Cref{fig:page1:b} shows the detection probability estimate we present in \Cref{eqn:geometry_binning_prob}.
Here, the probability that an object of a specified size (in our case a radius of 40~cm, similar in size to a dog) will be detected if placed at a certain location.
A screenshot of a single simulation time-step is super-imposed in grayscale to contextualize the scale of the visualization.
At close range, where occlusions from other objects are rare, this plot strongly reflects the simplified geometry of \Cref{fig:related:geo}, while still accurately capturing self-occlusions.

\begin{figure}[!t]
    \centering
    \includegraphics[width=\linewidth]{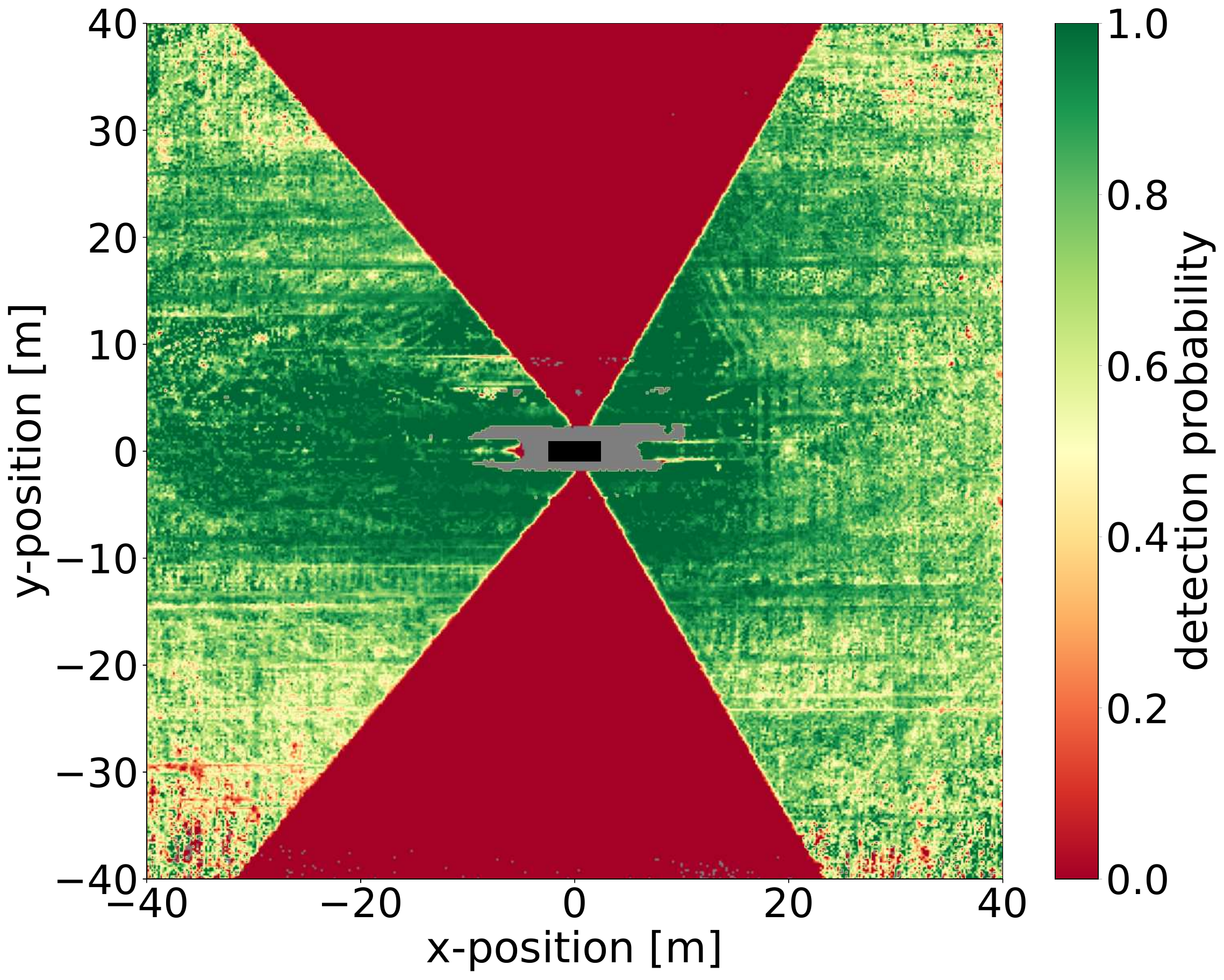}
    \caption{Detection probability of a small object at obstacle height for the camera setup illustrated in \Cref{fig:page1:a}.}
    \label{fig:detection_prob}
\end{figure}

At further ranges and typical obstacle height \mbox{($z\in [0.5~m,\,2.0~m]$)}, \Cref{fig:detection_prob} shows the gradual decay of detection probability as with increasing distance due to limited sensor resolution and occlusion through other traffic participants.
The following two experiments investigate and disentangle the effects of these two phenomena at the example of two LiDAR setups.

\subsection{Comparing visibility for different LiDAR positions}
\label{sec:experiments:lidar_pos}
\begin{figure}[!t]
    \includegraphics[width=1.0\linewidth]{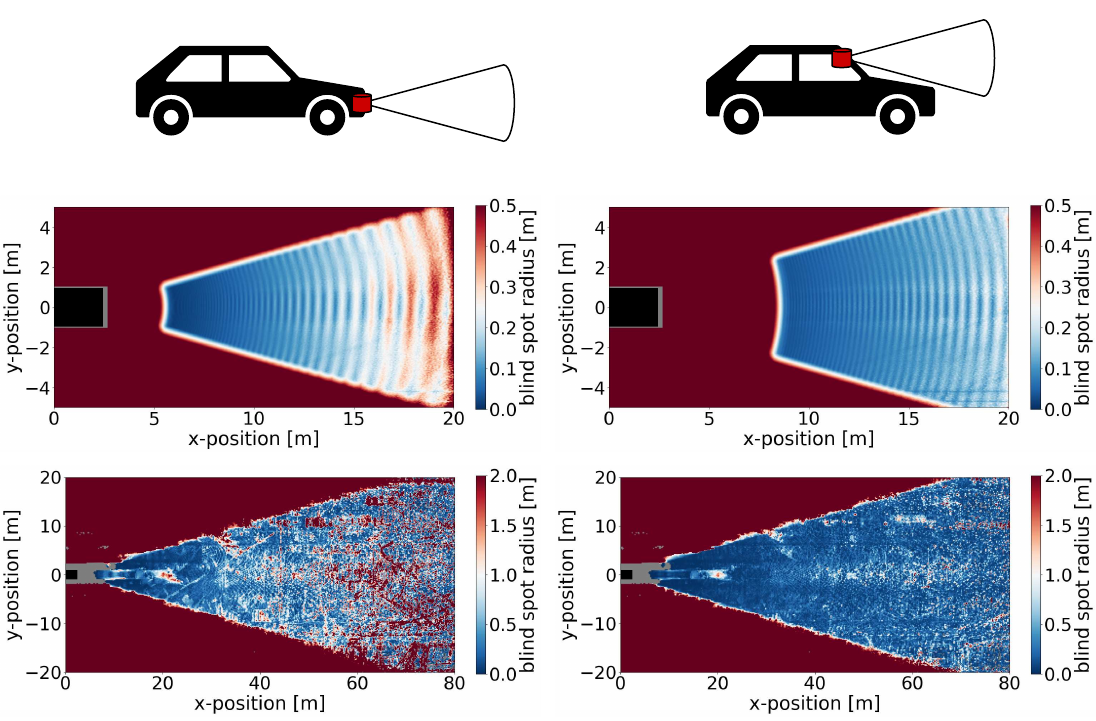}
    \caption{Comparing two common LiDAR positions for highway applications.
    The estimated blind spot size is shown in bird's-eye-view.
    The plots in the center show a close-up at ground level (z~$\approx$~0~m).
    The bottom shows medium range at a typical obstacle height (z $\in$ [0.5,\,2] m).
    Both sensors have identical parameters (128 channels) apart from their mounting position.}%
    \label{fig:would_you_see_a_child}%
\end{figure}

The second use case we illustrate in this section is the comparison of two different mounting positions for the same sensor.
In this case, we examine the effect of the sensor placement on the coverage of the sensor, both illustrated in \Cref{fig:would_you_see_a_child} (top) and numerically compared in \Cref{tab:sensor_positions}.
The first mounting position (\Cref{fig:would_you_see_a_child} top left) is a typical central grille-mounted position, prominently used for RADAR sensors in Mercedes-Benz vehicles among other manufacturers, but also commonly used for LiDAR sensors, such as in some Audi A7 models \cite{gitlin2019AudiA72019}.
The second mounting position we compare is a roof-mounted sensor position at the top of the vehicle's windscreen.
This position is prominently featured in NIO vehicles such as the EC6 \cite{NIOAssistedIntelligent}, or the Volvo EX90 \cite{klaymanVolvoElectricXC902021}.

As seen in \Cref{fig:would_you_see_a_child} (center, left vs right), the roof-mounted sensor position has a much shorter blind spot at ground level in front of the hood before the first LiDAR rays intersect the ground level.
As seen in \Cref{tab:sensor_positions} (row \emph{close range (20 m) ground}), this smaller blind spot close to the ego vehicle is reflected accurately in a smaller aggregated blind spot radius metric of 1.41~m, compared to 1.93~m for the roof-mounted LiDAR.
As many of the rays from the grille-mounted LiDAR sensor immediately hit the ground, the visibility at ground level begins to diminish rapidly, and gaps in the sensor's coverage already start to increase significantly at approximately 18 to 20 meters in front of the vehicle (see \Cref{fig:would_you_see_a_child}, center left).
One can also see that the region of coverage at similar distance is slightly wider for the roof-mounted sensor than for the grille-mounted position.

\begin{table}[b]
\centering
\resizebox{\columnwidth}{!}{%
\begin{tabular}{@{}lrr@{}}
\toprule%
\textbf{Sensor Position} & \multicolumn{1}{r}{\textbf{Front Grille LiDAR}} & \multicolumn{1}{r}{\textbf{Front Roof LiDAR}} \\
\midrule%
\emph{Region of Interest} &\multicolumn{2}{r}{\textbf{mean detection probability [\%] \textuparrow}}\\
\midrule%
\textit{close range (20~m) ground}     & 43.51  & \textbf{49.81} \\
\textit{medium range (80~m) ground}    & 16.38  & \textbf{22.62} \\
\textit{long range (160~m) ground}     & 9.73   & \textbf{12.82} \\
\midrule%
\textit{close range (20~m) obstacles}  & 49.93  & \textbf{59.05} \\
\textit{medium range (80~m) obstacles} & 42.08  & \textbf{49.05} \\
\textit{long range (160~m) obstacles}  & 25.73  & \textbf{30.69} \\
\midrule \midrule%
\emph{Region of Interest} & \multicolumn{2}{r}{\textbf{mean blind spot radius [m] \textdownarrow}} \\
\midrule%
\textit{close range (20~m) ground}     & \textbf{1.41}   & 1.93  \\
\textit{medium range (80~m) ground}    & 4.57   & \textbf{3.78}  \\
\textit{long range (160~m) ground}     & 11.38  & \textbf{10.29} \\ \midrule
\textit{close range (20~m) obstacles}  & 2.76   & \textbf{2.25}  \\
\textit{medium range (80~m) obstacles} & 6.66   & \textbf{5.74}  \\
\textit{long range (160~m) obstacles}  & 13.15  & \textbf{11.84} \\
\bottomrule%
\end{tabular}%
}%
\caption{Comparing the two LiDAR setups illustrated in \Cref{fig:would_you_see_a_child} at different ranges using our two presented metrics.}
\label{tab:sensor_positions}
\end{table}

When comparing the blind spot radius at obstacle height at further range (\Cref{fig:would_you_see_a_child}, bottom), a different phenomenon becomes clear:
Since the ego vehicle in our scenario is often following other vehicles in urban traffic, the view behind the ego vehicle is often obstructed by the vehicle being followed.
This results in a noticeable blind spot behind the occluding vehicle at approximately 22 meters in front of the ego vehicle for both setups.
However, due to the lower mounting position of the grille-mounted sensor, this blind spot is much larger and more pronounced.
Similarly, the lower mounting position results in more frequent occlusion by other objects in traffic, which generally results in much less consistent coverage at further range.
This is also reflected in \Cref{tab:sensor_positions}, where the detection probability declines much faster with increased range, and  the mean blind spot radius also increases faster with increasing range.

\subsection{Comparing different LiDAR sensor resolutions}
\label{sec:experiments:lidar_res}
The third use case we compare is the impact of different lidar sensor resolutions on a setup with a fixed mounting position.
For this experiment, we use the same roof-mounted setup as described in \Cref{sec:experiments:lidar_pos}, and illustrated in \Cref{fig:would_you_see_a_child} (top right).
As mechanical LiDAR sensors are often sold with varying vertical resolution (often referred to as the number of \emph{channels}), we inspect the result of changing the vertical sensor resolution while keeping the mounting position as well as the viewing frustum unchanged.
We compare a sensor model with 32, 64, and 128 channels, which are common denominations sold by LiDAR manufacturers.
\Cref{tab:sensor_resolutions} shows the aggregated metrics provided by our method.
As can be seen in this table, at close range, the sensor resolution does not have a large impact on detecting obstacles, whereas the point density with which the ground is sampled is greatly reduced.
At further distance, this reduction in sampling density is much more pronounced, and leads to a rapid decline in detection performance at increasing range for the low-resolution 32-channel sensor.
Correlated with the decline in detection performance, the mean blind spot radius also rises at further range for all tested sensor resolutions.

\begin{table}[t]
\centering
\resizebox{\columnwidth}{!}{%
\begin{tabular}{@{}lrrr@{}}
\toprule%
\textbf{Sensor resolution} & \multicolumn{1}{r}{\textbf{32 channels}} & \multicolumn{1}{r}{\textbf{64 channels}} & \multicolumn{1}{r}{\textbf{128 channels}} \\
\midrule%
\emph{Region of Interest} &\multicolumn{3}{r}{\textbf{mean detection probability [\%] \textuparrow}}\\
\midrule%
\textit{close range (20~m) ground}     &  27.96  &  42.88  & \textbf{49.81} \\
\textit{medium range (80~m) ground}    &   9.98  &  15.82  & \textbf{22.62} \\
\textit{long range (160~m) ground}     &   4.79  &  8.74  & \textbf{12.82} \\
\midrule%
\textit{close range (20~m) obstacles}  &  57.47  &  58.58  & \textbf{59.05} \\
\textit{medium range (80~m) obstacles} &  36.01  &  46.44  & \textbf{49.05} \\
\textit{long range (160~m) obstacles}  &  19.00  &  28.38  & \textbf{30.69} \\
\midrule \midrule%
\emph{Region of Interest} & \multicolumn{3}{r}{\textbf{mean blind spot radius [m] \textdownarrow}} \\
\midrule%
\textit{close range (20~m) ground}     &   2.11  &  1.98  & \textbf{1.93}  \\
\textit{medium range (80~m) ground}    &   5.06  &  4.21  & \textbf{3.78}  \\
\textit{long range (160~m) ground}     &  12.02  &  10.85  & \textbf{10.29} \\ \midrule
\textit{close range (20~m) obstacles}  &   2.28  &  2.27  & \textbf{2.25}  \\
\textit{medium range (80~m) obstacles} &   6.07  &  5.85  & \textbf{5.74}  \\
\textit{long range (160~m) obstacles}  &  12.28  &  12.00  & \textbf{11.84} \\
\bottomrule%
\end{tabular}%
}%
\caption{Comparing the effect of different vertical resolutions, i.e. the number of channels for the 'Front Roof' LiDAR setup illustrated in \Cref{fig:would_you_see_a_child} (right).}
\label{tab:sensor_resolutions}
\end{table}

%% file: sections/03_impl_details.tex
\section{Implementation Details \todo{0.5 pages}}
\label{sec:impl-details}
The method we presented as described in \Cref{sec:methods} is agnostic to the underlying simulation engine.
In this section, we detail our concrete implementation choices in order to simplify the reproduction of our method and results.
Our code is made available at \href{https://github.com/Pyrestone/CanYouSeeMeNow}{\url{github.com/Pyrestone/CanYouSeeMeNow}}

\subsection{Reference sensor setup}
\label{sec:impl:ref_setup}
To emulate a best-case-scenario sensor setup, we randomly sample reference sensor positions from a shell around the ego vehicle's bounding box.
At each time-step, the reference sensor is assigned a new position and orientation relative to the ego vehicle.
To compute this shell, we enlarge the ego vehicle's bounding box by 0.5~m upwards as well as in the horizontal directions, and subtract the ego vehicle's bounding box to receive the final sampling volume.
For maximum coverage, we use a simulated reference sensor with 1024 channels and 1024 points per channel.
The channels are evenly distributed between 0 and -90° elevation angle, and span 360° azimuth angle.
The orientation of the reference sensor is sampled uniformly from yaw values in the range \mbox{[-180°,\,180°]}, and pitch and roll values in the range \mbox{[-45°,\,45°].}

\subsection{Regions of Interest}
\label{sec:impl:roi}
In the various tables in this paper, we list coverage metrics for different regions of interest.
This means that the metric value is aggregated across the grid within the entire region of interest.
For the forward-facing setups evaluated in \Cref{tab:sensor_positions,tab:sensor_resolutions}, this region of interest ranges from x-coordinates zero to the maximum range stated in the table (e.g. $x \in [0~m,\,20~m]$ for the region of interests denoted "close range (20~m)").
The sideways extent of the region-of interest is chosen centrally with an aspect ratio of 2:1, as seen in \cref{fig:would_you_see_a_child}, e.g. $y \in [-5~m,\,5~m]$ for the region of interests denoted "close range (20~m)".
For all plots and metrics in this paper, we bin the vertical axis in two regions: ground level, where points lie in the range $z \in [-0.5,0.5]$, and obstacle height, which we define as $z \in [0.5,2.0]$.
The slight deviation in ground height is necessary, as the ground height varies slightly in our simulation environment (e.g. for sidewalks vs roads), and the roll and pitch motion of the ego vehicle also results in slight changes of the vertical position of the ground level in the vehicle coordinate system.

\subsection{Performance Optimizations}
\label{sec:impl:performance}
Due to the large number of points from both the reference sensor and the simulated sensor setup, a naive pair-wise minimum search across all combinations of reference and sensor points in \cref{eqn:distance_metric} becomes computationally very expensive.
To accelerate this step, we implement a python library based on a CUDA-acelerated KD-Tree nearest neighbor implementation by \cite{waldGPUfriendlyParallelAlmost2022a}, for which we provide a python package at \href{https://github.com/Pyrestone/numpy_cukd}{\url{github.com/Pyrestone/numpy_cukd}}.
Based on our benchmarks, this is currently the fastest public KD-Tree implementation available for this purpose in python.

%% file: sections/05_conclusion.tex
\section{Conclusion \todo{0.25 pages}}
\label{sec:conclusion}
In this work, we presented a simple but elegant method for analyzing the blind spots and coverage of LiDAR and camera sensor setups for autonomous or highly-automated vehicles, and similar mobile robotics applications.
By unifying both cameras and LiDAR sensors into a unified point cloud interface, our method is able show blind spots of both camera and LiDAR setups with one or multiple sensors.
By leveraging high-fidelity simulation data, our presented method also enables deeper investigations into the effects of varying sensor positions and resulting occlusions, or sensor attributes such as varying sensor resolution.
Our method is simple to implement, agnostic to the choice of simulation environment, and provides valuable visualizations of coverage and blind spots, as well as comparable metrics of mean blind spot radius and detection probability across a defined region of interest.
We show the versatility of our method in three experiments, which capture deep insights into the effect of different degrees of freedom in sensor setup design.